\pdfoutput=1

\documentclass[11pt]{article}

\usepackage[]{acl}

\usepackage{times}
\usepackage{latexsym}

\usepackage[utf8]{inputenc} 
\usepackage[T1]{fontenc}    
\usepackage{hyperref}       
\usepackage{url}            
\usepackage{booktabs}       
\usepackage{amsfonts}       
\usepackage{nicefrac}       
\usepackage{microtype}      
\usepackage{xcolor}         
\usepackage{bm,bbm,dsfont}
\usepackage{comment}
\usepackage{pgfplots}

\usepackage{pifont}
\usepackage{amsfonts}
\usepackage{multicol,multirow}
\usepackage{amsmath, amssymb, amsthm}
\usepackage[ruled]{algorithm2e}

\usepackage{bm,bbm,dsfont}

\usepackage{xspace,mfirstuc,tabulary}
\usepackage{enumitem}
\usepackage{makecell}
\usepackage{array}
\usepackage{cancel}
\usepackage{subfigure}

\usepackage{tikz-dependency}
\pgfplotsset{width=7cm,compat=1.13}

\usepackage[T1]{fontenc}
\usepackage[utf8]{inputenc}

\usepackage{microtype}

\title{Dependency Parsing as MRC-based Span-Span  Prediction}
\author{Leilei Gan$^{\blacklozenge}$, 
Yuxian Meng$^{\clubsuit}$, Kun Kuang$^\blacklozenge$, Xiaofei Sun$^{\blacklozenge\clubsuit}$ \\
{\bf  
Chun Fan$^{\spadesuit}$, 
Fei Wu$^\blacklozenge$ and Jiwei Li$^{\blacklozenge\clubsuit}$}\\
  $^\blacklozenge$Zhejiang University, $^\clubsuit$Shannon.AI, $^\spadesuit$ Computer Center, Peking University\\
  $^\spadesuit$Peng Cheng Laboratory, $^\spadesuit$National Biomedical Imaging Center, Peking University\\
  \{leileigan, kunkuang, wufei, jiwei\_li\}@zju.edu.cn \\
  \{yuxian\_meng, xiaofei\_sun\}@shannonai.com, fanchun@pku.edu.cn
}

\begin{document}
\maketitle

\begin{abstract}
Higher-order methods for dependency parsing can partially but not fully address the issue that  edges in dependency trees should be constructed at the text span/subtree level rather than word level. In this paper, we propose a new  method for  dependency parsing  to  address this issue. The proposed method constructs dependency trees by directly modeling span-span (in other words, subtree-subtree) relations. It consists of two modules: the {\it text span proposal module} which proposes candidate text spans, each of which represents a subtree in the dependency tree denoted by (root, start, end); and the {\it span linking module}, which constructs links between proposed spans. We use the machine reading comprehension (MRC) framework as the backbone to formalize  the span linking module, where one span is used as query to extract the text span/subtree it should be linked to. The proposed method has the following merits: (1)  it addresses the  fundamental problem that edges in a dependency tree should be constructed between  subtrees; (2) the MRC framework allows the method to retrieve missing spans in the span proposal stage, which leads to higher recall for eligible spans. Extensive experiments on the PTB, CTB and Universal Dependencies (UD) benchmarks demonstrate the effectiveness of the proposed method.
\footnote{Accepted by ACL2022}
\footnote{The code is available at \url{https://github.com/ShannonAI/mrc-for-dependency-parsing}}
\end{abstract}

\section{Introduction}
Dependency parsing is a basic and fundamental task in natural language processing (NLP) \citep{eisner2000bilexical,nivre2003efficient,mcdonald2005non}. 
Among existing efforts for dependency parsers, graph-based models 
\cite{mcdonald2005online,pei-etal-2015-effective} are a widely used category of models, which
cast the task as 
finding the optimal maximum spanning tree in the directed graph.
Graph-based models provide 
 a more global view than shift-reduce models \cite{zhang-nivre-2011-transition,chen2014fast},
 leading to better performances. 

\begin{figure}
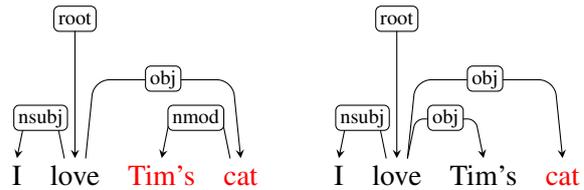

    \begin{minipage}[t]{0.45\linewidth}
    \centering
    \begin{dependency}
    \begin{deptext}[column sep=0.2cm]
    I \& love \& {\color{red}Tim's} \& {\color{red}cat} \\
    \end{deptext}
    \depedge{2}{1}{nsubj}
    \depedge{4}{3}{nmod}
    \depedge{2}{4}{obj}
    \deproot{2}{root}
    \end{dependency}
  \end{minipage}
  \hfill
  \begin{minipage}[t]{0.45\linewidth}
    \centering
    \begin{dependency}
    \begin{deptext}[column sep=0.2cm]
   I \& love \& Tim's \& {\color{red}cat} \\
    \end{deptext}
    \depedge{2}{1}{nsubj}
    \depedge{2}{3}{obj}
    \depedge{2}{4}{obj}
    \deproot{2}{root}
    \end{dependency}
  \end{minipage}%
  \caption{Two possible dependency trees for sentence ``I love Tim's cat''. 
For the tree on the right hand side, if we look at the token-token level,   `love" being linked to ``cat" is correct.
But at the subtree-subtree level, the linking is incorrect since the span/subtree behind ``cat" is incorrect.  }
    \label{dep}
\end{figure}

Graph-based methods are faced with a  challenge: they construct dependency edges by using word pairs as basic units for modeling, which is insufficient because  dependency parsing performs at the span/subtree level. For example, 
Figure \ref{dep} shows
two possible dependency trees for the sentence ``I love Tim's cat''. 
In both cases, at the token level, ``love" is linked to ``cat". 
If we only consider token-token relations,  the second case  of `love" being linked to ``cat" is correct.
But if we view the tree at the subtree-subtree level, the linking is incorrect since the span/subtree behind ``cat" is incorrect. 
   Although higher-order methods are able to alleviate this issue by aggregating information across adjacent edges,  they can not fully address the issue.
In nature, 
the token-token strategy can be viewed as a coarse  simplification of  the span-span (subtree-subtree) strategy, where the root token in
the token-token strategy
 can be viewed as the {\bf average} of all spans covering it.
 We would like an approach that directly models span-span relations using {\bf exact} subtrees behind tokens, rather than the {average} of all spans covering it.

To address this challenge, in this work, we propose a  model for dependency parsing that 
directly operates at the
span-span relation level. The proposed model consists of two modules: (1) the {\it text span proposal module} which  proposes eligible candidate text spans, each of which represents a subtree in the dependency tree denoted by (root,  start,  end);
(2) and the {\it span linking module}, which constructs links between proposed spans to form the final dependency tree.
We use the machine reading comprehension (MRC) framework as the backbone to formalize the span linking module in an MRC setup, where one span is used as a query to extract the text span/subtree it should be linked to.
In this way, the proposed model is able to directly model span-span relations and build the complete dependency tree in a bottom-up recursive manner.

The proposed model provides benefits in the following three aspects: (1)
firstly, 
it naturally addresses the shortcoming of token-token modeling in vanilla graph-based approaches and directly performs at the span level; (2) with the MRC framework, the left-out spans in the span proposal stage can still be retrieved at the span linking stage, and thus the negative effect of unextracted spans can be alleviated; and (3) the MRC formalization allows us to take advantage of existing state-of-the-art MRC models, with which the model expressivity can be enhanced, leading to better performances.

We are able to achieve new SOTA performances on PTB, CTB and UD benchmarks, which demonstrate the effectiveness of the proposed method.

\section{Related Work}
Transition-based dependency parsing incrementally constructs a dependency tree from input words through a sequence of shift-reduce actions \citep{zhang-nivre-2011-transition,chen2014fast,zhou2015neural,dyer2015transition,yuan2019bidirectional,han-etal-2019-enhancing,mohammadshahi-henderson-2020-graph}.
Graph-based dependency parsing searches through the space of all possible dependency trees for a tree that maximizes a specific  score
\citep{pei-etal-2015-effective,wang-etal-2018-improved,zhang-etal-2019-empirical}. 

Graph-based dependency parsing is first introduced by \citet{mcdonald2005online,mcdonald2005non}. They formalized the task of dependency parsing as finding the maximum spanning tree (MST) in directed graphs and used the large margin objective \citep{crammer2003} to efficiently train the model. 
\citet{zhang-etal-2016-probabilistic} introduced a probabilistic convolutional neural network (CNN) for graph-based dependency parsing to model third-order dependency information
\citet{wang-chang-2016-graph,kiperwasser-goldberg-2016-simple} proposed to employ LSTMs as an encoder to extract features, which are then used to \text{score} dependencies between words.
\citet{zhang2015high,zhang-etal-2019-empirical,wang-tu-2020-second} integrated higher-order features across adjacent dependency edges to build the dependency tree.
\citet{ji-etal-2019-graph} captured high-order dependency information by using graph neural networks. 
The biaffine approach \citep{dozat2016deep} is a particular kind of graph-based method improving upon vanilla scoring functions in graph-based dependency parsing. 
\citet{ma-etal-2018-stack} combined biaffine classifiers and pointer networks to build dependency trees in a top-down manner. 
\citet{jia-etal-2020-semi,zhang-etal-2020-efficient} extended the biaffine approach to the conditional random field (CRF) framework. 
\citet{mrini-etal-2020-rethinking} incorporated label information into the self-attention structure \citep{vaswani2017attention} for biaffine dependency parsing.

\section{Method}

\subsection{Notations}
Given a sequence of input tokens $\mathbf{s}=(w_0, w_1, ...,\\ w_n)$, where $n$ denotes the length of the sentence and $w_0$ is a dummy token representing the root of the sentence, we formalize the task of dependency parsing as finding the tree with the highest \text{score} among all possible trees rooted at $w_0$ .
\begin{equation}\label{argmax_tree}
    \hat{T}=\arg\max_{T_{w_0}}\text{score}(T_{w_0})
\end{equation}
Each token $w_i$ in the input sentence corresponds to a subtree $T_{w_i}$ rooted at $w_i$
within in the full tree $T$, and the subtree 
can be characterized by a text span, with the index of its leftmost token being $T_{w_i}.s$ in the original sequence, and 
the index of its rightmost token being $T_{w_i}.e$ in the original sequence. 
As shown in the first example of Figure \ref{dep}, the span covered by
the subtree
 $T_\text{love}$ is the full sentence ``I love Tim's cat'', and the span covered by the subtree $T_\text{cat}$ is ``Tim's cat''.
Each directional arc $w_i \rightarrow w_j$ in $T$ represents a parent-child relation
between $T_{w_i}$ and $T_{w_j}$, 
 where 
 $T_{w_j}$ is a subtree of $T_{w_i}$. This implies  that the text span covered by $T_{w_j}$ is fully contained by the text span covered by $T_{w_i}$.

It is worth noting that the currently proposed paradigm can only handle the projective situation. 
We will get back to how to adjust the current paradigm to non-projective situation in Section \ref{Inference}.

\subsection{Scoring Function}
With notations defined in the previous section, we now illustrate how to compute the $\text{score}(T_{w_0})$ in Eq.(\ref{argmax_tree}).
Since we want to model the span-span relations inside a dependency tree, where the tree is composed by spans and the links
between them, we formalize the scoring function as:
\begin{equation}
\begin{aligned}
\text{score}&(T_{w_0}) = \sum_{i=1}^{n}\text{score}_\text{span}(T_{w_i}) \\
&+ \lambda \sum_{(w_i \rightarrow w_j) \in T_{w_0}}{\text{score}_\text{link}(T_{w_i}, T_{w_j})}
\label{tree_score}
\end{aligned}
\end{equation}
where $\text{score}_\text{span}(T_{w_i})$ represents how likely the subtree rooted at $w_i$ covers the text span from $T.s$ to $T.e$. $\text{score}_\text{link}(T_{w_i}, T_{w_j})$ represents how likely tree $T_{w_j}$ is a subtree of $T_{w_i}$, i.e. there is an arc from $w_i$ to $w_j$, and $\lambda$ is a hyper-parameter to balance $\text{score}_\text{span}$ and $\text{score}_\text{link}$.
We will illustrate the details how to compute $\text{score}_\text{span}(T)$ and $\text{score}_\text{link}(T_1, T_2)$ in the following sections. Table \ref{tab:example} shows all the spans and links for the left tree in Figure \ref{dep}.

\begin{table}[t]
    \centering
    \small
    \begin{tabular}{ll}\toprule 
    \multicolumn{2}{c}{Sentence: {\it root I love Tim's cat}}\\
       {\bf Spans}  &  {\bf Links} \\\midrule
       $T_\text{root}$: root I love Tim's cat  & $T_\text{root}\to T_\text{love}$\\
       $T_\text{I}$: I  & $T_\text{love}\to T_\text{I}$\\
       $T_\text{love}$: I love Tim's cat  & $T_\text{love}\to T_\text{cat}$\\
       $T_\text{Tim's}$: Tim's  & $T_\text{cat}\to T_\text{Tim's}$\\
       $T_\text{cat}$: Tim's cat  &\\
    \bottomrule
    \end{tabular}
    \caption{Spans and links for the left tree in Figure \ref{dep}.}
    \label{tab:example}
\end{table}

\subsection{Span Proposal Module}
In this section, we introduce the span proposal module. This module gives each tree $T_{w_i}$ a score $\text{score}_\text{span}(T_{w_i})$ in Eq.(\ref{tree_score}), which represents how likely the subtree rooted at $w_i$ covers the text span from $T_{w_i}.s$ to $T_{w_i}.e$.
The score can be decomposed into two components -- the score for the left half span from  $w_i$ to $T_{w_i}.s$, and the score for the right half span from $w_i$ to $T_{w_i}.e$, given by:
\begin{equation}
\begin{aligned}
\text{score}_\text{span}(T_{w_i}) &= \text{score}_\text{start}(T_{w_i}.s|w_i)  \\
&+ \text{score}_\text{end}(T_{w_i}.e|w_i)
\end{aligned}
\label{span_score}
\end{equation}
We propose to formalize  $\text{score}_\text{start}(T_{w_i}.s|w_i)$ 
as the score for  the text span starting  at $T_{w_i}.s$, ending at $w_i$, by transforming the task to a text span extraction problem. 
Concretely, we use the biaffine function 
\citep{dozat2016deep} to score the text span by computing $\text{score}_\text{start}(j|i)$ -- the score of the tree rooted at at ${w_i}$ and staring at $w_j$:
\begin{equation}
\text{score}_\text{start}(j|i)=\mathbf{x}_i^\top U_{\text{start}}\mathbf{x}_j+\mathbf{w}_{\text{start}}^\top \mathbf{x}_j
\label{score_start}
\end{equation}
where $U\in\mathbb{R}^{d \times d}$ and $\mathbf{w}\in\mathbb{R}^{d}$ are trainable parameters, $\mathbf{x}_i\in\mathbb{R}^{d}$ and $\mathbf{x}_j\in\mathbb{R}^{d}$ are
token representations of $w_i$ and $w_j$ respectively.
To obtain $\mathbf{x}_i$ and $\mathbf{x}_j$, we pass the sentence $s$ to pretrained models such as BERT \cite{devlin2018bert}.
$\mathbf{x}_i$ and $\mathbf{x}_j$ are the last-layer representations output from BERT for $w_i$ and $w_j$.
We use the following loss to optimize the left-half span proposal module:
\begin{multline}
\mathcal{L}_\text{span}^{\text{start}} = - \sum_{i=1}^{n} \text{log} \frac{\exp(\text{score}_\text{start}(T_{w_i}.s|i))}{\sum_{j=1}^{n}\exp(\text{score}_\text{start}(j|i))}
\label{span_start_loss}
\end{multline}
This objective enforces the model to find the correct span start $T_{w_i}.s$ for each word $w_i$. We ignore loss for $w_0$, the dummy root token.

$\text{score}_{\text{end}}(T_{w_i}.e|w_i)$ can be computed in the similar way, where the model extracts the text span rooted at index $w_i$ and ending at $T_{w_i}.e$:
\begin{equation}
\text{score}_\text{end}(j|i)=\mathbf{x}_i^\top U_{\text{end}}\mathbf{x}_j+\mathbf{w}_{\text{end}}^\top \mathbf{x}_j
\label{score_end}
\end{equation}
The loss to optimize the right-half span proposal module:
\begin{multline}
\mathcal{L}_\text{span}^{\text{end}} = - \sum_{i=1}^{n} \text{log}\frac{\exp(\text{score}_\text{end}(T_{w_i}.e|i))}{\sum_{j=1}^{n}\exp(\text{score}_\text{end}(j|i))} 
\label{span_end_loss}
\end{multline}
Using the left-half span score  in Eq.(\ref{score_start}) and the right-half span score in Eq.(\ref{score_end}) to compute 
the full span score in
Eq.(\ref{span_score}), we are able to compute the score for any subtree, with 
 text span starting at $T_{w_i}.s$, ending at $T_{w_i}.e$ and rooted at $w_i$. 

\subsection{Span Linking Module}
Given two subtrees $T_{w_i}$ and $T_{w_j}$, the span linking module gives a \text{score} -- $\text{score}_\text{link}(T_{w_i}, T_{w_j})$ to represent the probability of $T_{w_j}$ being a subtree of $T_{w_i}$.
This means that $T_{w_i}$ is the parent of $T_{w_j}$, and that the span associated with $T_{w_j}$, i.e., $(T_{w_j}.s, T_{w_j}.e)$
is fully contained in the  span associated with 
 $T_{w_i}$, i.e., $(T_{w_i}.s, T_{w_i}.e)$. 


We propose to use the machine reading comprehension framework as the backbone to compute this score.  It operates on the triplet \{context  ($X$), query ($q$) and answer ($a$)\}. 
The context $X$ is the original sentence $\mathbf{s}$. The query $q$ is the child span $(T_{w_j}.s, T_{w_j}.e)$. 
And we wish to extract the answer, which is the parent span  $(T_{w_i}.s, T_{w_i}.e)$ from the context input sentence $s$.
The basic idea here is that using the child span to query the full sentence gives direct cues for identifying the corresponding parent span, and this is more effective than simply feeding two extracted spans and then determining whether they have the parent-child relation.

\paragraph{Constructing Query} Regarding the query, we should  consider both the span and its root. The query is thus formalized as follows:
\begin{multline}
\texttt{<sos>}, T_{w_j}.s, T_{w_j}.s+1, ..., T_{w_j}-1, \texttt{<sor>}, \\
T_{w_j}, \texttt{<eor>}, T_{w_j}+1,..., \\
 T_{w_j}.e-1, T_{w_j}.e, \texttt{<eos>} 
\end{multline}
where \texttt{<sos>}, \texttt{<sor>}, \texttt{<eor>}, and \texttt{<eos>} are special tokens, which respectively denote the start of span, the start of root, the end of root, and the end of span. 
One issue with the way above to construct query is that the position information of $T_{w_j}$ is not included in the query.
In practice, we turn to a more convenient strategy where the query is the original sentence, with special tokens
 \texttt{<sos>}, \texttt{<sor>}, \texttt{<eor>}, and \texttt{<eos>} used to denote the position of the child. In this way, position information for child $T_{w_j}$ can be naturally considered.  

\paragraph{Answer Extraction} 
The answer is the parent, with the span $T_{w_i}.s, T_{w_i}.e$ rooted at $T_{w_i}$. 
We can directly take the framework from the MRC model by identifying the start and end of the answer span, respectively denoted by $\text{score}_\text{parent}^{s}(T_{w_i}.s|T_{w_j})$
and $\text{score}_\text{parent}^{e}(T_{w_i}.e|T_{w_j})$.
We also wish to identify the root $T_{w_i}$ from the answer, which is characterized by the score of $w_i$ being the root of the span, denoted by 
 $\text{score}_\text{parent}^{r}(w_i|T_{w_j})$.
 Furthermore, since we also want to identify the relation category between the  parent and the child, the score signifying the relation label $l$ is needed to be added, which is denoted by 
 $\text{score}_\text{parent}^l(l|T_{w_j}, w_i)$.

For quadruple 
 $(T_{w_i}.s, T_{w_i}.e, T_{w_j}, l)$, which denotes the span $T_{w_i}.s, T_{w_i}.e$ rooted at $w_i$, 
 the final score for it
 being the answer to $T_{w_j}$, and the relation between the subtrees is $l$, is given by:
\begin{equation}
\begin{aligned}
&\text{score}_\text{parent}(T_{w_i}|T_{w_j})=\\
& \text{score}_\text{parent}^r(w_i|T_{w_j})  + \text{score}_\text{parent}^s(T_{w_i}.s|T_{w_j})+  \\
&\text{score}_\text{parent}^e(T_{w_i}.e|T_{w_j}) + \text{score}_\text{parent}^l(l|T_{w_j}, w_i)
\label{link_forward_score}
\end{aligned}
\end{equation}
In the MRC setup, the input is the concatenation of the query and the context, denoted by $\{\texttt{<cls>}, \text{query},\texttt{<sep>}, \text{context}\}$, where $\texttt{<cls>}$ and $\texttt{<sep>}$ are special tokens. The input is fed to BERT, and we 
 obtain   representations
for each input token. Let $\mathbf{h}_t$ denote the representation for the token with index $t$ output from BERT. 
The probability of $t^\text{th}$ token being the root of the answer, which is denoted by 
$\text{score}_\text{parent}^{r}(w_t|T_{w_j})$ is the softmax function over all constituent tokens in the context:  
\begin{equation}
    \text{score}_\text{parent}^r(w_t|T_{w_j})=\frac{\exp{(\mathbf{h}^\top_\text{root} \times \mathbf{h}_t)}}{\sum_{t'\in\text{context}} \exp(\mathbf{h}^\top_\text{root} \times \mathbf{h}_{t'})}
\end{equation}
where $\mathbf{h}^\top_\text{root}$ is trainable parameter. $\text{score}_\text{parent}^{s}$ and $\text{score}_\text{parent}^{e}$ can be computed in the similar way:
\begin{equation}
\begin{aligned}
    &\text{score}_\text{parent}^s(w_t|T_{w_j})=\frac{\exp{(\mathbf{h}^\top_\text{start} \times \mathbf{h}_t)}}{\sum_{t'\in\text{context}} \exp(\mathbf{h}^\top_\text{start} \times \mathbf{h}_{t'})} \\
    &\text{score}_\text{parent}^e(w_t|T_{w_j})=\frac{\exp{(\mathbf{h}^\top_\text{end} \times \mathbf{h}_t)}}{\sum_{t'\in\text{context}} \exp(\mathbf{h}^\top_\text{end} \times \mathbf{h}_{t'})}
 \end{aligned}
\end{equation}
 For $ \text{score}_\text{parent}^l(l|T_{w_j}, w_i)$, which denotes the relation label between $T_{w_i}$ and $T_{w_j}$, we can compute it in a simple way.
 Since $\mathbf{h}_{w_i}$ already encodes information for $\mathbf{h}_{w_j}$ through self-attentions, 
  the representation $\mathbf{h}_{w_i}$ for $w_i$
is directly fed to the softmax function over all labels in the label set $\mathcal{L}$:
\begin{equation}
   \text{score}_\text{parent}^l(l|T_{w_j}, w_i) =\frac{\exp{(\mathbf{h}^\top_{l} \times \mathbf{h}_{w_i})}}{\sum_{l'\in \mathcal{L}} \exp(\mathbf{h}^\top_{l'} \times \mathbf{h}_{w_i})}
\end{equation}

 
\paragraph{Mutual Dependency}
A closer look at Eq.(\ref{link_forward_score}) reveals that it only models
the uni-directional dependency relation that $T_{w_i}$ is the parent of $T_{w_j}$.
This is suboptimal since if $T_{w_i}$ is a parent answer of $T_{w_j}$, $T_{w_j}$ should be a child answer of $T_{w_i}$.
We thus propose to use $T_{w_i}$ as the query $q$ and $T_{w_j}$ as the answer $a$.
\begin{equation}
\begin{aligned}
&\text{score}_\text{child}(T_{w_j}|T_{w_i})=\\
& \text{score}_\text{child}^r(w_j|T_{w_i})  + \text{score}_\text{child}^s(T_{w_j}.s|T_{w_i})+  \\
&\text{score}_\text{child}^e(T_{w_j}.e|T_{w_i}) + \text{score}_\text{child}^l(l|T_{w_i}, T_{w_j})
\label{link_backward_score}
\end{aligned}
\end{equation}
The final \text{score} $\text{score}_\text{link}$ is thus given by:
\begin{equation}
\begin{aligned}
\text{score}_\text{link}(T_{w_i},T_{w_j}) &= \text{score}_\text{child}(T_{w_j} | T_{w_i}) \\
 &+ \text{score}_\text{parent}(T_{w_i}|w_j)
\end{aligned}
\label{linking_score}
\end{equation}
Since one tree may have multiple children but can only have one parent, we use the multi-label cross entropy loss $\mathcal{L}_\text{parent}$ for $\text{score}_\text{parent}(T_{w_i}|T_{w_j})$ and use the binary cross entropy loss $\mathcal{L}_\text{child}$ for $\text{score}_\text{child}(T_{w_j}|T_{w_i})$. 
We jointly optimize these two losses $\mathcal{L}_\text{link}=\mathcal{L}_\text{parent}+\mathcal{L}_\text{child}$ for  span linking.

\subsection{Inference}
\label{Inference}
Given an input sentence $\mathbf{s}=(w_0, w_1, w_2, ..., w_n)$, the number of all possible subtree spans $(w_i, T_{w_i}.s, T_{w_i}.e)$ is $\mathcal{O}(n^3)$, 
and therefore running MRC procedure for every candidate span is  computationally prohibitive. 
A naive solution is to use the span proposal module to extract top-k scored spans rooted at 
each token. This gives rise to a set of span candidates $\mathcal{T}$ with size $1+n \times k$ (the root token $w_0$ produces only one span), where each candidate span is associated with its subtree span score $\text{score}_\text{span}(\cdot)$. 
Then we construct 
the optimal dependency tree based only on these extracted spans by linking them. 
This strategy obtains a local optimum 
for Eq.(\ref{tree_score}), 
because we want to compute the optimal solution for  the first  part $ \sum_{i=1}^{n}\text{score}_\text{span}(T_{w_i})$ depending on the second part of Eq.(\ref{tree_score}), i.e., $\sum_{(w_i \rightarrow w_j) \in T_{w_0}}\text{score}_\text{link}(T_{w_i}, T_{w_j})$.  But in this naive strategy, the second part  is computed after the first part. 

It is worth noting that the naive solution of using only the top-k scored spans has another severe issue: spans left out
at the span proposal stage can never be a part of the final prediction, since the span linking module only operates on the proposed spans. This would not be a big issue if top-k is large enough to recall almost every span in ground-truth. 
However, span proposal is intrinsically harder than span linking because the span proposal module lacks the triplet span information that is used by the span linking module. Therefore, we propose to use the span linking module to retrieve more correct spans. 
Concretely, for every span $T_{w_j}$ proposed by the span proposal module, we use $\arg\max\text{score}_\text{parent}(T_{w_i}|T_{w_j})$ to retrieve its parent with the highest score as additional span candidates.
Recall that span proposal proposed $1+n \times k$ spans. Added by spans proposed by  the span linking module, the maximum number of candidate spans is $1+ 2\times n \times k$.
The MRC formalization behind the span linking module improves the recall rate as  missed spans at the span proposal stage can still be retrieved at this stage.

\begin{algorithm}[t]
    \footnotesize
      \SetKwInOut{Input}{Input}
      \SetKwInOut{Output}{Output}
      \Input{Input sentence $\mathbf{s}$, span candidates $\mathcal{T}$, 
             span scores $\text{score}_\text{span}(T), \forall T\in\mathcal{T}$}
      \Output{Highest score of every span $\text{score}(T), \forall T \in \mathcal{T}$}

      {\color{blue}{
      \tcc*[h]{Compute linking scores based on Eq.(\ref{linking_score})}}}
      
      $\text{score}_\text{link}(T_1,T_2)\leftarrow \text{score}_\text{parent}(T_1|T_2) +  \text{score}_\text{child}(T_2|T_1),\forall (T_1,T_2)\in\mathcal{T}$

      {\color{blue}{
      \tcc*[h]{Compute $\text{score}(T), \forall T\in\mathcal{T}$}}}
    
      \For{$len \leftarrow 0$ \KwTo $n$}{
          \For{$T \leftarrow \mathcal{T}$}{
               \If{$T.e-T.s = len$}{

                {\color{blue}{
                \tcc*[h]{$T$ covers a single word}}}

                    \eIf{$len = 0$}{ 
                        $\text{score}(T) \leftarrow \text{score}_\text{span}(T)$
                    }{

                        {\color{blue}{
                        \tcc*[h]{$\mathcal{C}$ is a set of direct subtrees composing $T$}}}
                        
                        $\text{score}(T)\leftarrow \text{score}_\text{span}(T) + \max_{\mathcal{C}(T)}(\sum_{T_i \in \mathcal{C}}[\text{score}(T_i) + \lambda \text{score}_\text{link}(T, T_i)])$
                    }
               }
          }
      }
      \caption{Projective Inference}
      \label{algo}
\end{algorithm}

\paragraph{Projective Decoding}
Given retrieved spans harvested in the proposal stage, 
we use a CKY-style bottom-up dynamic programming algorithm to find the projective tree with the highest score based on Eq.(\ref{tree_score}). The algorithm is present in Algorithm \ref{algo}.
The key idea is  that we can generalize the definition of $\text{score}(T_{w_0})$ in Eq.(\ref{tree_score}) to any $w$ by the following definition:
\begin{equation}
\begin{aligned}
\text{score}&(T_{w}) = \sum_{T_{w_i} \subseteq T_{w}} \text{score}_\text{span}(T_{w_i}) \\
&+ \lambda \sum_{(w_i \rightarrow w_j) \in T_{w}}{\text{score}_\text{link}(T_{w_i}, T_{w_j})}
\label{any_tree_score}
\end{aligned}
\end{equation}
where $\{T_{w_i} \mid T_{w_i} \subseteq T_{w}, i=0,1,...,n\}$ is all subtrees inside $T_w$, i.e. there is a path in $T_w$ like
$$w \rightarrow w_{i_1} \rightarrow ..., \rightarrow w_{i}$$
Using this definition, we can rewrite $\text{score}(T_w)$ in a recursive manner:
\begin{equation}
\begin{aligned}
&\text{score}(T_w) = \text{score}_\text{span}(T_w) \\
&+ \sum_{T_{w_j} \in \mathcal{C}(T_w)}[\text{score}(T_{w_j}) + \lambda~\text{score}_\text{link}(T_w, T_{w_j})]
\label{recursive_tree_score}
\end{aligned}
\end{equation}
where $\mathcal{C}(T_w)=\{ T_{w_i} \mid (w \rightarrow w_i) \in T_w, i=0,1,...n \}$ is the set of
all direct subtrees of $T_w$.

\paragraph{Non-Projective Decoding}
It is noteworthy that effectively finding a set of subtrees composing a tree $T$ requires trees to be projective (the projective property guarantees every subtree is a continuous span in text), and experiments in Section \ref{experiments} show that this algorithm performs well on datasets
where most trees are projective, but performs worse when a number of trees are non-projective. 
To address
this issue, we adapt the proposed strategy to the MST (Maximum Spanning Tree) algorithm \citep{mcdonald2005non}. 
The key point of MST is to obtain the score for each pair of tokens $w_i$ and $w_j$  (rather than spans) , denoted by $\text{score}_\text{edge}(w_i, w_j)$.
We propose that the score to link $w_i$ and $w_j$  is the highest score achieved by two spans  respectively rooted at $w_i$ and $w_j$: 
\begin{equation}
\begin{aligned}
&\text{score}_\text{edge}(w_i, w_j) = \max_{T_{w_i}, T_{w_j}} [\text{score}_\text{span}(T_{w_i})\\& + \text{score}_\text{span}(T_{w_j}) + \lambda \text{score}_\text{link}(T_{w_i}, T_{w_j})]
\label{span_loss}
\end{aligned}
\end{equation}
The final score for tree $T$ is given by:
\begin{equation}
\text{score}(T) = \sum_{(w_i \rightarrow w_j) \in T} \text{score}_\text{edge}(w_i, w_j) 
\end{equation}
Here, MST can be readily used for decoding.

\section{Experiments} 
\label{experiments}
\subsection{Datasets and Metrics}
We carry out experiments on three widely used dependency parsing benchmarks: the English Penn Treebank v3.0 (PTB) dataset \citep{marcus-etal-1993-building}, the Chinese Treebank v5.1 (CTB) dataset \citep{xue-etal-2002-building} and the Universal Dependency Treebanks v2.2 (UD) \citep{nivre-etal-2016-universal} where we select 12 languages for evaluation. 
We follow \citet{ma-etal-2018-stack} to process all datasets. The PTB dataset contains 39832 sentences for training and 2416 sentences for test. The CTB dataset contains 16091 sentences for training and 1910 sentences for test. The statistics for 12 languages in UD dataset are the same with \citet{ma-etal-2018-stack}.
We use the  unlabeled attachment score (UAS) and labeled attachment score (LAS) for evaluation. Punctuations are ignored in all datasets during evaluation.

\subsection{Experiment Settings}
We compare the proposed model to the following baselines: (1) {\bf Biaffine}, (2) {\bf StackPTR}, (3){\bf GNN}, (4) {\bf MP2O}, (5) {\bf CVT}, (6) {\bf LRPTR}, (7) {\bf HiePTR}, (8) {\bf TreeCRF}, (9) {\bf HPSG}, (10) {\bf HPSG+LA}, (11) {\bf MulPTR}, (12) {\bf SynTr}. The details of these baselines are left to the supplementary materials due to page limitation.
We group experiments into three categories: without pretrained models, with BERT and with RoBERTa. 
To implement a span-prediction parsing model without pretrained models, 
we use the QAnet \cite{yu2018qanet} for span prediction. 
To enable apple-to-apple comparisons, 
we implement our proposed model, the Biaffine model, MP2O \cite{wang-tu-2020-second}
 based on 
BERT$_\text{large}$ \cite{devlin2018bert} and RoBERTa$_\text{large}$ \citep{yinhan2019roberta} for PTB, 
BERT and 
RoBERTa-wwm$_\text{large}$ \citep{cui2019pre} for CTB,
BERTBase-Multilingual-Cased and XLM-RoBERTa$_\text{large}$ for UD. 
We apply both projective decoding and non-projective MST decoding for all datasets.
For all experiments, we concatenate 100d POS tag embedding with 1024d pretrained token embeddings, then project them to 1024d using a linear layer. 
Following \citet{mrini-etal-2020-rethinking}, 
we further add 1-3 additional encoder layers on top  to let POS  embeddings well interact with pretrained token embeddings.  
POS tags are predicted using the Stanford NLP package \cite{manning2014stanford}. 
We tried two different types of additional encoders: Bi-LSTM \citep{hochreiter1997long} and Transformer \citep{vaswani2017transformer}. For Bi-LSTM, the number of hidden size is 1024d. For Transformer, the number of attention heads and hidden size remain the same as pretrained models (16 for attention heads and 1024d for hidden size).
We use 0.1 dropout rate for pretrained models and 0.3 dropout rate for additional layers.
We use Adam \citep{kingma2014adam} as optimizer. 
The weight parameter $\lambda$ is tuned on the development set. The code is implemented by PyTorch 1.6.0 and MindSpore.


\subsection{Main Results}
Table \ref{tab:ptb_ctb} compares our model to existing state-of-the-art models on PTB/CTB test sets. 
As can be seen, for models without pretrained LM, the proposed span-prediction model based on QAnet outperforms all baselines, illustrating the effectiveness of the proposed span-prediction framework for dependency parsing. 
For BERT-based models, the proposed span-prediction models outperform Biaffine model based on BERT, along with other competitive baselines.
On PTB, performances already outperform all previous baselines, except on the LAS metric in comparison to HiePTR (95.46 vs. 95.47)  on PTB, but underperform
RoBERTa-based models. 
On CTB, the proposed span-prediction model obtains a new SOTA performance of 
93.14\% UAS.
For RoBERTa-based models,  
 the proposed model 
  achieves a new SOTA performance  of 97.24\% UAS and 95.49\% LAS on PTB. 
     As PTB and CTB contain almost only projective trees, the projective decoding strategy significantly outperforms the non-projective MST algorithm.
It is worth noting that, since MulPTR, HPSG and HPSG+LA
 rely on additional labeled data of constituency parsing,  results for HPSG are not comparable to ours. We list them here for reference purposes.  

Table \ref{tab:ud} compares our model with existing state-of-the-art methods on UD test sets. 
Other than es, where the proposed model slightly underperforms the SOTA model by 0.02, the proposed model enhanced with XLM-RoBERTa achieves SOTA performances on all other 11 languages, with an average performance boost of 0.3.
As many languages in UD have a notable portion of non-projective trees,  MST decoding significantly outperforms projective decoding, leading to new SOTA performances in almost all language sets.

\begin{table}[t]
    \centering 
    \small
    \scalebox{0.9}{
    \begin{tabular}{lcccc}
        \toprule 
        & \multicolumn{2}{c}{{\bf PTB}} & \multicolumn{2}{c}{{\bf CTB}}\\
        & UAS & LAS & UAS & LAS \\
        \midrule 
                 \multicolumn{5}{c}{\textit{with additional labelled constituency parsing data}}\\
        {\it MulPTR$^\flat$} & 96.06 &  94.50 & 90.61  &89.51 \\
        {\it MulPTR+BERT$^\flat$}&96.91 &95.35& 92.58& 91.42\\
        {\it HPSG$^\flat$} & 97.20 & 95.72 & - & - \\
        {\it HPSG+LA$^\flat$} & 97.42 & 96.26 & 94.56 & 89.28 \\
       \hline 

                \multicolumn{5}{c}{\underline{without Pretrained Models}}\\
         {\it Biaffine}& 95.74 & 94.08 & 89.30 & 88.23 \\
         {\it StackPTR} & 95.87 & 94.19 & 90.59 & 89.29 \\
                {\it GNN} & 95.87 & 94.15 & 90.78 & 89.50 \\
	{\it LRPTR} &96.04 & 94.43 & - & - \\
	{\it HiePTR} &96.18& 94.59 &90.76 &{89.67}\\
	{\it TreeCRF} & 96.14& 94.49 & - & - \\
	{\it Ours-Proj} & {\bf 96.42} & {\bf 94.71}& {\bf 91.15} & {\bf 89.68}\\
	  {\it Ours-Nproj}& 96.33 & 94.60 &90.12&89.55\\
        \hline 
        \multicolumn{5}{c}{\underline{with Pretrained Models}}\\
        \multicolumn{5}{c}{\textit{with BERT}}\\
       {\it Biaffine} & 96.78 & 95.29 & 92.58 & 90.70 \\
       {\it MP2O} & 96.91 & 95.34 & 92.55 & {90.69} \\
       {\it SynTr+RNGTr} & 96.66 & 95.01 & {92.98} & 91.18\\
       {\it HiePTR} & 97.05 &{\bf 95.47}& 92.70 &{\bf 91.50}\\
      {\it Ours-Proj} & {\bf 97.18} & {95.46} &{\bf 93.14} & {91.27} \\
       {\it Ours-Nproj} & 97.09 & 95.35 & 93.06 & 91.21 \\
       \multicolumn{5}{c}{\textit{with RoBERTa}}\\
              {\it Biaffine} & 96.87 & 95.34 & 92.45 & 90.48 \\
               {\it MP2O}& 96.94 & 95.37 & 92.37 & 90.40\\
       {\it Ours-Proj} & \textbf{97.24} & \textbf{95.49} & {\bf 92.68} & {\bf 90.91} \\
       {\it Ours-Nproj} & 97.14 & 95.39 & 92.58 & 90.83 \\
       \bottomrule 
    \end{tabular}
    }
    \caption{Results for different models on PTB and CTB. $^\flat$: These approaches utilized both dependency
    and constituency information in their approach, thus is not comparable to ours.}
    \label{tab:ptb_ctb}
\end{table}

\begin{table*}[t]
    \centering 
    \small 
    \scalebox{0.75}{
    \begin{tabular}{lcccccccccccccc}
        \toprule 
        & bg & ca & cs & de & en & es & fr & it & nl & no & ro & ru & Avg. \\
       projective\% & 99.8 & 99.6 & 99.2 & 97.7 & 99.6 & 99.6 & 99.7 & 99.8 & 99.4 & 99.3 & 99.4 & 99.2 & 99.4 \\ 
        \midrule 
       {\it GNN} & 90.33 & 92.39 & 90.95 & 79.73 & 88.43 & 91.56 & 87.23 & 92.44 & 88.57 & 89.38 & 85.26 &91.20 & 89.37 \\
       \hline 
       \multicolumn{14}{c}{\underline{+BERT}}\\
       \citep{sun2020cross} & 90.88 & 93.67& 91.52 & 86.64 & 90.60 & 93.01 & 90.65 & 94.05 & 92.81 & 91.05 & 86.59 & 92.35 & 91.15\\
       {\it MP2O} & 91.30 & 93.60 & 92.09 & 82.00 & 90.75 & 92.62 & 89.32 & 93.66 & 91.21 & 91.74 & 86.40 & 92.61 & 91.02\\
       {\it Biaffine} & 92.72 &93.88 & 92.70 & 83.65 & 91.31 & 91.89 & 90.87 & 94.02 & 92.24 & 93.50 & 88.11 & 94.37 & 91.61\\
       {\it Ours-Proj} & 93.42 & 93.85 & 92.90 & 84.88 & 91.74 & 92.05 & 91.50 & {94.62} & 92.42 & 93.98 & 88.52 & 94.50 & 92.03\\ 
        {\it Ours-NProj} & {93.61} & {94.22} & {93.48} & {85.14} & {91.77} & {92.50} & {91.52} & 94.60 & {92.82} & {94.24} & 88.48 & {94.73} & 92.30\\
       \multicolumn{14}{c}{\underline{+XLM-RoBERTa}}\\
              {\it MP2O} & 91.42 & 93.75 & 92.15 & 82.20 & 90.91 & 92.60 & 89.51 & 93.79 & 91.45 & 91.95 & 86.50 & 92.81 & 90.75\\
       {\it Biaffine} & 93.04 & 94.15 & 93.57 & 84.84 & 91.93 & \textbf{92.64} & 91.64 & 94.07 & 92.78 & 94.17 & 88.66 & 94.91 &92.15\\
       {\it Ours-Proj} & 93.61 & 94.04 & 93.1 & 84.97 & 91.92 & 92.32 & 91.69 & \textbf{94.86} & 92.51 & 94.07 & \textbf{88.76} & 94.66 & 92.21\\ 
       & (+0.57) & (-0.11) & (-0.47) & (+0.13) & (-0.01) & (-0.32) & (+0.05) & \textbf{(+0.79)} & (-0.27) & (-0.10) & \textbf{(+0.10)} & (-0.25) & (+0.06) \\ 
        {\it Ours-NProj} & \textbf{93.76} & \textbf{94.38} & \textbf{93.72} & \textbf{85.23} & \textbf{91.95} & {92.62} & \textbf{91.76} & 94.79 & \textbf{92.97} & \textbf{94.50} & 88.67 & \textbf{95.00} & \textbf{92.45} \\ 
         & \textbf{(+0.72)} & \textbf{(+0.23)} & \textbf{(+0.15)} & \textbf{(+0.39)} & \textbf{(+0.02)} & (-0.02) & \textbf{(+0.12)} & (+0.72) & \textbf{(+0.19)} & \textbf{(+0.33)} & (+0.01) & \textbf{(+0.09)} & (\textbf{+0.30}) \\ 
       \bottomrule 
    \end{tabular}
    }
    \caption{LAS for different model on UDv2.2. We use ISO 639-1 codes
to represent languages from UD.}
    \label{tab:ud}
\end{table*}

\section{Ablation Study and Analysis}
We use  PTB to understand behaviors of the proposed model. As projective decoding works best for PTB, scores reported in this section are all from projective decoding.

\subsection{Effect of Candidate Span Number}
We would like to study the effect of the number of candidate spans proposed by the span proposal module, i.e., the value of $k$. We vary the value of $k$ from 1 to 25.
As shown in Table \ref{tab:ablation_k}, increasing values of $k$ leads to higher UAS, and the performance stops increasing once $k$ is large enough ($k > 15$).
More interestingly, even though $k$ is set to 1, which means that only one candidate span is proposed for each word, the final UAS score is 96.94, a score that is very close to the best result 97.24 and surpasses most existing methods as shown in Table \ref{tab:ptb_ctb}. 
These results verify that the proposed approach can accurately extract and link the dependency spans.

\begin{table}
\centering
\small
\scalebox{0.9}{
    \begin{tabular}{lcccccc}
        \toprule
        $k$ & 1 & 2 & 5 & 10 & 15 & 20 \\
        \midrule
        UAS & 96.94 & 97.10 & 97.22 & 97.23 & 97.24 & 97.23\\
    \bottomrule
    \end{tabular}
    }
    \caption{Effect of number of span candidates}
    \label{tab:ablation_k}
\end{table}

\begin{table}
    \centering
    \small
    \scalebox{0.9}{
    \begin{tabular}{lccccc}
        \toprule
        $k$ & 1 & 2 & 5 & 10 & 15 \\
        \midrule
        recall w/o link & 97.09 & 98.88 & 99.57 & 99.77 & 99.86 \\
        recall w/ link & 97.76 & 99.14 & 99.69 & 99.85 & 99.92 \\  
    \bottomrule
    \end{tabular}
    }
    \caption{Span recall with/without  span linking module}
    \label{tab:ablation_backward}
\end{table}
\begin{table}[t]
    \centering
    \small
    \scalebox{0.9}{
    \begin{tabular}{lcc}
        \toprule 
        Model & UAS & LAS \\
        \midrule
         Full & 97.24 & 95.49  \\
         token(query)-token(answer) & 96.67(-0.57) & 95.15(-0.34)  \\
        span(query)-token(answer) & 97.17(-0.07) & 95.44(-0.05)  \\
        token(query)-span(answer)& 97.11(-0.13) & 95.41(-0.08)  \\
        -mutual & 97.18(-0.06) & 95.43(-0.06)  \\
        \bottomrule
    \end{tabular}
    }
    \caption{The effect of removing different parts from scoring function}
    \label{tab:ablation_score}
\end{table}
\subsection{Effect of Span Retrieval by Span Linking}
As shown in Table \ref{tab:ablation_backward}, 
span recall significantly improves with the presence of the span linking stage. 
This is in line with our expectation,  since spans missing at the proposal module can be retrieved by QA model in the span linking stage. 
Recall boost narrows down when $k$ becomes large, which is expected as more candidates are proposed at the proposal stage. 
The span linking stage can improve 
 computational efficiency by using a smaller number of proposed spans  while achieving the same performance.
\begin{figure}[t]
  \centering
  \begin{minipage}[t]{0.32\linewidth}
    \centering
    \includegraphics[width=1\linewidth]{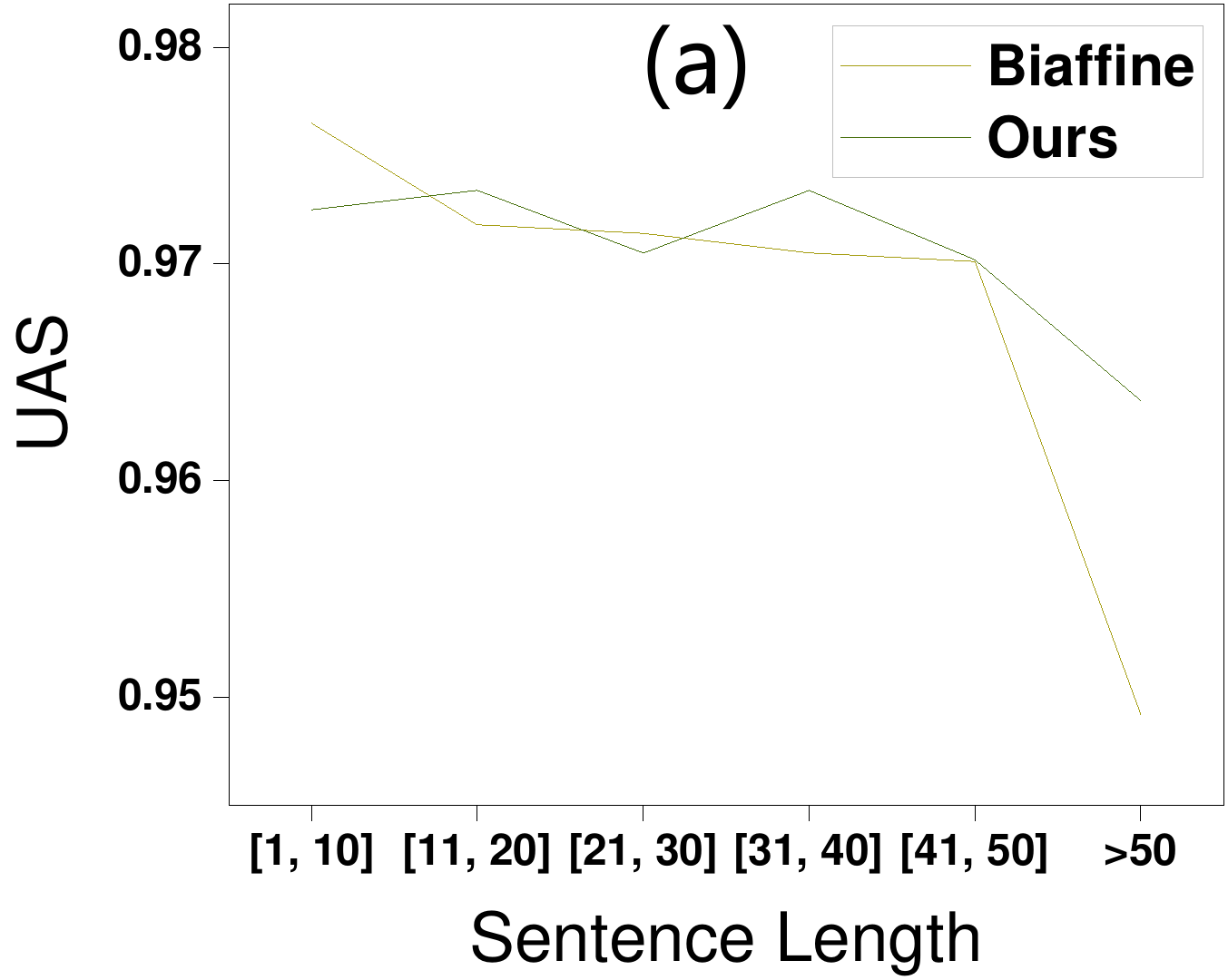}
  \end{minipage}%
  \hfill
  \begin{minipage}[t]{0.32\linewidth}
    \centering
    \includegraphics[width=1\linewidth]{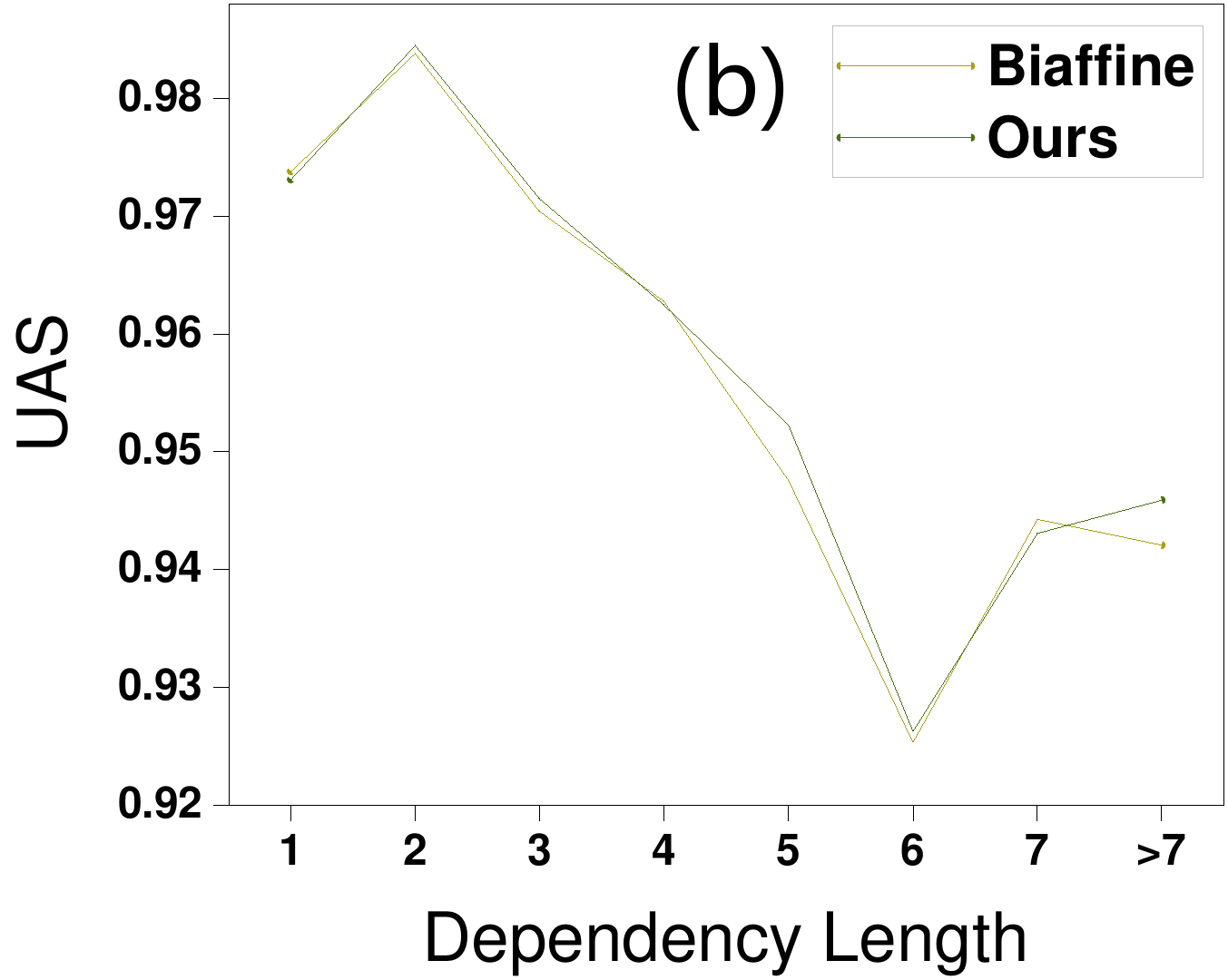}
  \end{minipage}%
  \hfill
  \begin{minipage}[t]{0.32\linewidth}
    \centering
    \includegraphics[width=1\linewidth]{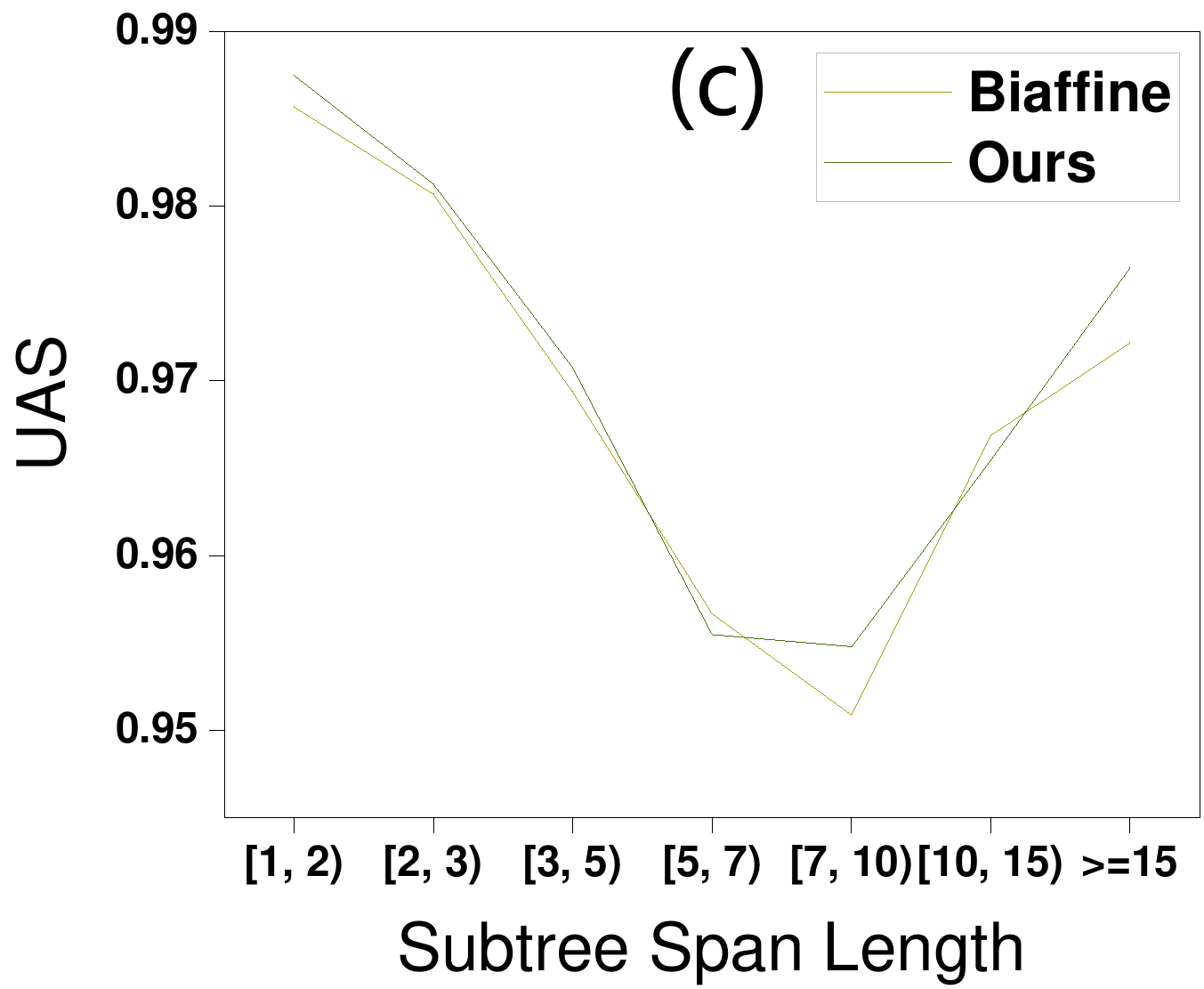}
  \end{minipage}%
  \caption{Performances on PTB test dev of Biaffine and our parser w.r.t. sentence length, dependency length and subtree span length.}
\label{fig:error_analysis}
\end{figure}

\subsection{Effect of Scoring Functions}
We study the effect of each part of the scoring functions used in the proposed model. 
Table \ref{tab:ablation_score} shows the results. We have the following observations:

\noindent
(1) {\bf token(query)-token(answer)}: we simplify the model by only signifying root token in  queries (child) and extract the root token in the context (parent). The model actually degenerates into a model similar to Biaffine by working at the token-token level. We observe significant performance decreases, 0.57 in UAS and 0.34 in LAS. 

\noindent
(2)  {\bf token(query)-span(answer)}: signifying  only token in queries (child) and extracting span in answers (parent)  leads to a decrease of 0.13 and 0.08 respectively for UAS and LAS. 

\noindent
(3)  {\bf span(query)-token(answer)}: signifying  spans in queries (child) but only extracting token in answers  (parent)  leads to a decrease of 0.07 and 0.05 respectively for UAS and LAS. 
(1), (2) and (3) demonstrate the necessity of modeling span-span  rather than token-token relations in dependency parsing: replacing  span-based strategy with token-based strategy 
 for either parent or child progressively leads to performance decrease.  
 
\noindent (4) Removing the {\bf Mutual Dependency} module which only uses child $\rightarrow$ parent relation and ignores parent $\rightarrow$ child relation also leads to performance decrease. 

\subsection{Analysis}
Following \citet{ma-etal-2018-stack,ji-etal-2019-graph}, we analyze performances of the Biaffine parser and the proposed method with respect to sentence length, dependency length, and subtree span length.
Results are shown in Figure \ref{fig:error_analysis}. 

\noindent\textbf{Sentence Length.} As shown in Figure \ref{fig:error_analysis}(a), the proposed parser achieves better performances on long sentences compared with Biaffine. Specially, when sentence length is greater than 50, the performance of the Biaffine parser  decreases significantly, while the proposed  parser has a much smaller drop (from 0.97 to 0.964).

\noindent\textbf{Dependency Length.} Figure \ref{fig:error_analysis}(b) shows the results with respect to dependency length. The proposed parser  shows its advantages on long-range dependencies. We suppose span-level information is beneficial for long-range dependencies.

\noindent\textbf{Subtree Span Length.} We further conduct experiments on subtree span length. We divide the average lengths of the two spans in the span linking module into seven buckets. We suppose our parser should show advantages on long subtree span, and the results in Figure \ref{fig:error_analysis}(c) verify our conjecture.

In summary, the span-span strategy works significantly better than the token-token strategy, especially for long sequences. 
This explanation is as follows: 
the token-token strategy can be viewed as a coarse  simplification of  the span-span strategy, where the root token in
the token-token strategy
 can be viewed as the {\bf average} of all spans covering it, 
while  in  the span-span strategy, it represents the exact span, rather than the average. The deviation from the average is relatively small from the extract when sequences are short, but becomes larger as sequence length grows, since the number of spans covering the token exponentially grows with length. This makes the token-token strategy work significantly worse for long sequences. 

\section{Conclusion}
In this paper, we propose to construct dependency trees by directly modeling span-span instead of token-token relations.
 We use the machine reading comprehension framework  to formalize the span linking module, where one span is used as a query to extract the text span/subtree it should be linked to.
Extensive experiments on the PTB, CTB and UD benchmarks show the effectiveness of the proposed method. 

\section*{Acknowledgement}
This work is supported by the Science and Technology Innovation 2030 - “New Generation Artificial Intelligence” Major Project (No. 2021ZD0110201), the Key R \& D Projects of the Ministry of Science and Technology (2020YFC0832500) and MindSpore Open Fund.
We would like to thank anonymous reviewers for their comments and suggestions. 

\bibliographystyle{acl_natbib}
\bibliography{parsing}

\begin{thebibliography}{45}
\expandafter\ifx\csname natexlab\endcsname\relax\def\natexlab#1{#1}\fi

\bibitem[{Chen and Manning(2014)}]{chen2014fast}
Danqi Chen and Christopher~D Manning. 2014.
\newblock A fast and accurate dependency parser using neural networks.
\newblock In \emph{Proceedings of the 2014 conference on empirical methods in
  natural language processing (EMNLP)}, pages 740--750.

\bibitem[{Clark et~al.(2018)Clark, Luong, Manning, and
  Le}]{clark-etal-2018-semi}
Kevin Clark, Minh-Thang Luong, Christopher~D. Manning, and Quoc Le. 2018.
\newblock Semi-supervised sequence modeling with cross-view training.
\newblock In \emph{Proceedings of the 2018 Conference on Empirical Methods in
  Natural Language Processing}, pages 1914--1925, Brussels, Belgium.
  Association for Computational Linguistics.

\bibitem[{Crammer et~al.(2006)Crammer, Dekel, Keshet, Shalev-Shwartz, and
  Singer}]{crammer2003}
Koby Crammer, Ofer Dekel, Joseph Keshet, Shai Shalev-Shwartz, and Yoram Singer.
  2006.
\newblock Online passive-aggressive algorithms.
\newblock \emph{J. Mach. Learn. Res.}, 7:551–585.

\bibitem[{Cui et~al.(2019)Cui, Che, Liu, Qin, Yang, Wang, and Hu}]{cui2019pre}
Yiming Cui, Wanxiang Che, Ting Liu, Bing Qin, Ziqing Yang, Shijin Wang, and
  Guoping Hu. 2019.
\newblock Pre-training with whole word masking for chinese bert.
\newblock \emph{arXiv preprint arXiv:1906.08101}.

\bibitem[{Devlin et~al.(2018)Devlin, Chang, Lee, and
  Toutanova}]{devlin2018bert}
Jacob Devlin, Ming-Wei Chang, Kenton Lee, and Kristina Toutanova. 2018.
\newblock Bert: Pre-training of deep bidirectional transformers for language
  understanding.
\newblock \emph{arXiv preprint arXiv:1810.04805}.

\bibitem[{Dozat and Manning(2016)}]{dozat2016deep}
Timothy Dozat and Christopher~D Manning. 2016.
\newblock Deep biaffine attention for neural dependency parsing.
\newblock \emph{arXiv preprint arXiv:1611.01734}.

\bibitem[{Dyer et~al.(2015)Dyer, Ballesteros, Ling, Matthews, and
  Smith}]{dyer2015transition}
Chris Dyer, Miguel Ballesteros, Wang Ling, Austin Matthews, and Noah~A Smith.
  2015.
\newblock Transition-based dependency parsing with stack long short-term
  memory.
\newblock \emph{arXiv preprint arXiv:1505.08075}.

\bibitem[{Eisner(2000)}]{eisner2000bilexical}
Jason Eisner. 2000.
\newblock Bilexical grammars and their cubic-time parsing algorithms.
\newblock In \emph{Advances in probabilistic and other parsing technologies},
  pages 29--61. Springer.

\bibitem[{Fern{\'a}ndez-Gonz{\'a}lez and
  G{\'o}mez-Rodr{\'\i}guez(2019)}]{fernandez2019left}
Daniel Fern{\'a}ndez-Gonz{\'a}lez and Carlos G{\'o}mez-Rodr{\'\i}guez. 2019.
\newblock Left-to-right dependency parsing with pointer networks.
\newblock \emph{arXiv preprint arXiv:1903.08445}.

\bibitem[{Fern{\'a}ndez-Gonz{\'a}lez and
  G{\'o}mez-Rodr{\'\i}guez(2020)}]{fernandez2020multitask}
Daniel Fern{\'a}ndez-Gonz{\'a}lez and Carlos G{\'o}mez-Rodr{\'\i}guez. 2020.
\newblock Multitask pointer network for multi-representational parsing.
\newblock \emph{arXiv preprint arXiv:2009.09730}.

\bibitem[{Fern{\'a}ndez-Gonz{\'a}lez and
  G{\'o}mez-Rodr{\'\i}guez(2021)}]{fernandez2021dependency}
Daniel Fern{\'a}ndez-Gonz{\'a}lez and Carlos G{\'o}mez-Rodr{\'\i}guez. 2021.
\newblock Dependency parsing with bottom-up hierarchical pointer networks.
\newblock \emph{arXiv preprint arXiv:2105.09611}.

\bibitem[{Han et~al.(2019)Han, Jiang, and Tu}]{han-etal-2019-enhancing}
Wenjuan Han, Yong Jiang, and Kewei Tu. 2019.
\newblock Enhancing unsupervised generative dependency parser with contextual
  information.
\newblock In \emph{Proceedings of the 57th Annual Meeting of the Association
  for Computational Linguistics}, pages 5315--5325, Florence, Italy.
  Association for Computational Linguistics.

\bibitem[{Hochreiter and Schmidhuber(1997)}]{hochreiter1997long}
Sepp Hochreiter and J{\"u}rgen Schmidhuber. 1997.
\newblock Long short-term memory.
\newblock \emph{Neural computation}, 9(8):1735--1780.

\bibitem[{Ji et~al.(2019)Ji, Wu, and Lan}]{ji-etal-2019-graph}
Tao Ji, Yuanbin Wu, and Man Lan. 2019.
\newblock Graph-based dependency parsing with graph neural networks.
\newblock In \emph{Proceedings of the 57th Annual Meeting of the Association
  for Computational Linguistics}, pages 2475--2485, Florence, Italy.
  Association for Computational Linguistics.

\bibitem[{Jia et~al.(2020)Jia, Ma, Cai, and Tu}]{jia-etal-2020-semi}
Zixia Jia, Youmi Ma, Jiong Cai, and Kewei Tu. 2020.
\newblock Semi-supervised semantic dependency parsing using {CRF} autoencoders.
\newblock In \emph{Proceedings of the 58th Annual Meeting of the Association
  for Computational Linguistics}, pages 6795--6805, Online. Association for
  Computational Linguistics.

\bibitem[{Kingma and Ba(2014)}]{kingma2014adam}
Diederik~P Kingma and Jimmy Ba. 2014.
\newblock Adam: A method for stochastic optimization.
\newblock \emph{arXiv preprint arXiv:1412.6980}.

\bibitem[{Kiperwasser and Goldberg(2016)}]{kiperwasser-goldberg-2016-simple}
Eliyahu Kiperwasser and Yoav Goldberg. 2016.
\newblock Simple and accurate dependency parsing using bidirectional {LSTM}
  feature representations.
\newblock \emph{Transactions of the Association for Computational Linguistics},
  4:313--327.

\bibitem[{Liu et~al.(2019)Liu, Ott, Goyal, Du, Joshi, Chen, Levy, Lewis,
  Zettlemoyer, and Stoyanov}]{yinhan2019roberta}
Yinhan Liu, Myle Ott, Naman Goyal, Jingfei Du, Mandar Joshi, Danqi Chen, Omer
  Levy, Mike Lewis, Luke Zettlemoyer, and Veselin Stoyanov. 2019.
\newblock Roberta: A robustly optimized bert pretraining approach.
\newblock \emph{arXiv preprint arXiv:1907.11692}.

\bibitem[{Ma et~al.(2018)Ma, Hu, Liu, Peng, Neubig, and
  Hovy}]{ma-etal-2018-stack}
Xuezhe Ma, Zecong Hu, Jingzhou Liu, Nanyun Peng, Graham Neubig, and Eduard
  Hovy. 2018.
\newblock Stack-pointer networks for dependency parsing.
\newblock In \emph{Proceedings of the 56th Annual Meeting of the Association
  for Computational Linguistics (Volume 1: Long Papers)}, pages 1403--1414,
  Melbourne, Australia. Association for Computational Linguistics.

\bibitem[{Manning et~al.(2014)Manning, Surdeanu, Bauer, Finkel, Bethard, and
  McClosky}]{manning2014stanford}
Christopher~D Manning, Mihai Surdeanu, John Bauer, Jenny~Rose Finkel, Steven
  Bethard, and David McClosky. 2014.
\newblock The stanford corenlp natural language processing toolkit.
\newblock In \emph{Proceedings of 52nd annual meeting of the association for
  computational linguistics: system demonstrations}, pages 55--60.

\bibitem[{Marcus et~al.(1993)Marcus, Santorini, and
  Marcinkiewicz}]{marcus-etal-1993-building}
Mitchell~P. Marcus, Beatrice Santorini, and Mary~Ann Marcinkiewicz. 1993.
\newblock Building a large annotated corpus of {E}nglish: The {P}enn
  {T}reebank.
\newblock \emph{Computational Linguistics}, 19(2):313--330.

\bibitem[{McDonald et~al.(2005{\natexlab{a}})McDonald, Crammer, and
  Pereira}]{mcdonald2005online}
Ryan McDonald, Koby Crammer, and Fernando Pereira. 2005{\natexlab{a}}.
\newblock Online large-margin training of dependency parsers.
\newblock In \emph{Proceedings of the 43rd Annual Meeting of the Association
  for Computational Linguistics (ACL’05)}, pages 91--98.

\bibitem[{McDonald et~al.(2005{\natexlab{b}})McDonald, Pereira, Ribarov, and
  Hajic}]{mcdonald2005non}
Ryan McDonald, Fernando Pereira, Kiril Ribarov, and Jan Hajic.
  2005{\natexlab{b}}.
\newblock Non-projective dependency parsing using spanning tree algorithms.
\newblock In \emph{Proceedings of human language technology conference and
  conference on empirical methods in natural language processing}, pages
  523--530.

\bibitem[{Mohammadshahi and
  Henderson(2020{\natexlab{a}})}]{mohammadshahi-henderson-2020-graph}
Alireza Mohammadshahi and James Henderson. 2020{\natexlab{a}}.
\newblock Graph-to-graph transformer for transition-based dependency parsing.
\newblock In \emph{Findings of the Association for Computational Linguistics:
  EMNLP 2020}, pages 3278--3289, Online. Association for Computational
  Linguistics.

\bibitem[{Mohammadshahi and
  Henderson(2020{\natexlab{b}})}]{mohammadshahi2020recursive}
Alireza Mohammadshahi and James Henderson. 2020{\natexlab{b}}.
\newblock Recursive non-autoregressive graph-to-graph transformer for
  dependency parsing with iterative refinement.
\newblock \emph{arXiv preprint arXiv:2003.13118}.

\bibitem[{Mrini et~al.(2020)Mrini, Dernoncourt, Tran, Bui, Chang, and
  Nakashole}]{mrini-etal-2020-rethinking}
Khalil Mrini, Franck Dernoncourt, Quan~Hung Tran, Trung Bui, Walter Chang, and
  Ndapa Nakashole. 2020.
\newblock Rethinking self-attention: Towards interpretability in neural
  parsing.
\newblock In \emph{Findings of the Association for Computational Linguistics:
  EMNLP 2020}, pages 731--742, Online. Association for Computational
  Linguistics.

\bibitem[{Nivre(2003)}]{nivre2003efficient}
Joakim Nivre. 2003.
\newblock An efficient algorithm for projective dependency parsing.
\newblock In \emph{Proceedings of the Eighth International Conference on
  Parsing Technologies}, pages 149--160.

\bibitem[{Nivre et~al.(2016)Nivre, de~Marneffe, Ginter, Goldberg, Haji{\v{c}},
  Manning, McDonald, Petrov, Pyysalo, Silveira, Tsarfaty, and
  Zeman}]{nivre-etal-2016-universal}
Joakim Nivre, Marie-Catherine de~Marneffe, Filip Ginter, Yoav Goldberg, Jan
  Haji{\v{c}}, Christopher~D. Manning, Ryan McDonald, Slav Petrov, Sampo
  Pyysalo, Natalia Silveira, Reut Tsarfaty, and Daniel Zeman. 2016.
\newblock {U}niversal {D}ependencies v1: A multilingual treebank collection.
\newblock In \emph{Proceedings of the Tenth International Conference on
  Language Resources and Evaluation ({LREC}'16)}, pages 1659--1666,
  Portoro{\v{z}}, Slovenia. European Language Resources Association (ELRA).

\bibitem[{Pei et~al.(2015)Pei, Ge, and Chang}]{pei-etal-2015-effective}
Wenzhe Pei, Tao Ge, and Baobao Chang. 2015.
\newblock An effective neural network model for graph-based dependency parsing.
\newblock In \emph{Proceedings of the 53rd Annual Meeting of the Association
  for Computational Linguistics and the 7th International Joint Conference on
  Natural Language Processing (Volume 1: Long Papers)}, pages 313--322,
  Beijing, China. Association for Computational Linguistics.

\bibitem[{Sun et~al.(2020)Sun, Li, and Zhao}]{sun2020cross}
Kailai Sun, Zuchao Li, and Hai Zhao. 2020.
\newblock Cross-lingual universal dependency parsing only from one monolingual
  treebank.
\newblock \emph{arXiv preprint arXiv:2012.13163}.

\bibitem[{Vaswani et~al.(2017{\natexlab{a}})Vaswani, Shazeer, Parmar,
  Uszkoreit, Jones, Gomez, Kaiser, and Polosukhin}]{vaswani2017transformer}
Ashish Vaswani, Noam Shazeer, Niki Parmar, Jakob Uszkoreit, Llion Jones,
  Aidan~N Gomez, \L~ukasz Kaiser, and Illia Polosukhin. 2017{\natexlab{a}}.
\newblock Attention is all you need.
\newblock In I.~Guyon, U.~V. Luxburg, S.~Bengio, H.~Wallach, R.~Fergus,
  S.~Vishwanathan, and R.~Garnett, editors, \emph{Advances in Neural
  Information Processing Systems 30}, pages 5998--6008. Curran Associates, Inc.

\bibitem[{Vaswani et~al.(2017{\natexlab{b}})Vaswani, Shazeer, Parmar,
  Uszkoreit, Jones, Gomez, Kaiser, and Polosukhin}]{vaswani2017attention}
Ashish Vaswani, Noam Shazeer, Niki Parmar, Jakob Uszkoreit, Llion Jones,
  Aidan~N Gomez, {\L}ukasz Kaiser, and Illia Polosukhin. 2017{\natexlab{b}}.
\newblock Attention is all you need.
\newblock In \emph{Advances in neural information processing systems}, pages
  5998--6008.

\bibitem[{Wang and Chang(2016)}]{wang-chang-2016-graph}
Wenhui Wang and Baobao Chang. 2016.
\newblock Graph-based dependency parsing with bidirectional {LSTM}.
\newblock In \emph{Proceedings of the 54th Annual Meeting of the Association
  for Computational Linguistics (Volume 1: Long Papers)}, pages 2306--2315,
  Berlin, Germany. Association for Computational Linguistics.

\bibitem[{Wang et~al.(2018)Wang, Chang, and Mansur}]{wang-etal-2018-improved}
Wenhui Wang, Baobao Chang, and Mairgup Mansur. 2018.
\newblock Improved dependency parsing using implicit word connections learned
  from unlabeled data.
\newblock In \emph{Proceedings of the 2018 Conference on Empirical Methods in
  Natural Language Processing}, pages 2857--2863, Brussels, Belgium.
  Association for Computational Linguistics.

\bibitem[{Wang and Tu(2020)}]{wang-tu-2020-second}
Xinyu Wang and Kewei Tu. 2020.
\newblock Second-order neural dependency parsing with message passing and
  end-to-end training.
\newblock In \emph{Proceedings of the 1st Conference of the Asia-Pacific
  Chapter of the Association for Computational Linguistics and the 10th
  International Joint Conference on Natural Language Processing}, pages 93--99,
  Suzhou, China. Association for Computational Linguistics.

\bibitem[{Xue et~al.(2002)Xue, Chiou, and Palmer}]{xue-etal-2002-building}
Nianwen Xue, Fu-Dong Chiou, and Martha Palmer. 2002.
\newblock Building a large-scale annotated {C}hinese corpus.
\newblock In \emph{{COLING} 2002: The 19th International Conference on
  Computational Linguistics}.

\bibitem[{Yu et~al.(2018)Yu, Dohan, Luong, Zhao, Chen, Norouzi, and
  Le}]{yu2018qanet}
Adams~Wei Yu, David Dohan, Minh-Thang Luong, Rui Zhao, Kai Chen, Mohammad
  Norouzi, and Quoc~V Le. 2018.
\newblock Qanet: Combining local convolution with global self-attention for
  reading comprehension.
\newblock \emph{arXiv preprint arXiv:1804.09541}.

\bibitem[{Yuan et~al.(2019)Yuan, Jiang, and Tu}]{yuan2019bidirectional}
Yunzhe Yuan, Yong Jiang, and Kewei Tu. 2019.
\newblock Bidirectional transition-based dependency parsing.
\newblock In \emph{Proceedings of the AAAI Conference on Artificial
  Intelligence}, volume~33, pages 7434--7441.

\bibitem[{Zhang et~al.(2020)Zhang, Li, and Zhang}]{zhang-etal-2020-efficient}
Yu~Zhang, Zhenghua Li, and Min Zhang. 2020.
\newblock Efficient second-order {T}ree{CRF} for neural dependency parsing.
\newblock In \emph{Proceedings of the 58th Annual Meeting of the Association
  for Computational Linguistics}, pages 3295--3305, Online. Association for
  Computational Linguistics.

\bibitem[{Zhang and Nivre(2011)}]{zhang-nivre-2011-transition}
Yue Zhang and Joakim Nivre. 2011.
\newblock Transition-based dependency parsing with rich non-local features.
\newblock In \emph{Proceedings of the 49th Annual Meeting of the Association
  for Computational Linguistics: Human Language Technologies}, pages 188--193,
  Portland, Oregon, USA. Association for Computational Linguistics.

\bibitem[{Zhang et~al.(2019)Zhang, Ma, and Hovy}]{zhang-etal-2019-empirical}
Zhisong Zhang, Xuezhe Ma, and Eduard Hovy. 2019.
\newblock An empirical investigation of structured output modeling for
  graph-based neural dependency parsing.
\newblock In \emph{Proceedings of the 57th Annual Meeting of the Association
  for Computational Linguistics}, pages 5592--5598, Florence, Italy.
  Association for Computational Linguistics.

\bibitem[{Zhang and Zhao(2015)}]{zhang2015high}
Zhisong Zhang and Hai Zhao. 2015.
\newblock High-order graph-based neural dependency parsing.
\newblock In \emph{Proceedings of the 29th Pacific Asia Conference on Language,
  Information and Computation}, pages 114--123.

\bibitem[{Zhang et~al.(2016)Zhang, Zhao, and
  Qin}]{zhang-etal-2016-probabilistic}
Zhisong Zhang, Hai Zhao, and Lianhui Qin. 2016.
\newblock Probabilistic graph-based dependency parsing with convolutional
  neural network.
\newblock In \emph{Proceedings of the 54th Annual Meeting of the Association
  for Computational Linguistics (Volume 1: Long Papers)}, pages 1382--1392,
  Berlin, Germany. Association for Computational Linguistics.

\bibitem[{Zhou et~al.(2015)Zhou, Zhang, Huang, and Chen}]{zhou2015neural}
Hao Zhou, Yue Zhang, Shujian Huang, and Jiajun Chen. 2015.
\newblock A neural probabilistic structured-prediction model for
  transition-based dependency parsing.
\newblock In \emph{Proceedings of the 53rd Annual Meeting of the Association
  for Computational Linguistics and the 7th International Joint Conference on
  Natural Language Processing (Volume 1: Long Papers)}, pages 1213--1222.

\bibitem[{Zhou and Zhao(2019)}]{zhou-zhao-2019-head}
Junru Zhou and Hai Zhao. 2019.
\newblock {H}ead-{D}riven {P}hrase {S}tructure {G}rammar parsing on {P}enn
  {T}reebank.
\newblock In \emph{Proceedings of the 57th Annual Meeting of the Association
  for Computational Linguistics}, pages 2396--2408, Florence, Italy.
  Association for Computational Linguistics.

\end{thebibliography}
\newpage

\appendix 

\section{Baselines}
We use the following baselines:
\begin{itemize}[noitemsep]
    \item {\bf Biaffine}: \citet{dozat2016deep} fed pairs of words into a biaffine classifier to determine the dependency relations between them.
    \item {\bf StackPTR}: \citet{ma-etal-2018-stack} combined pointer network with transition-based method to make it benefits from the information of whole sentence and all previously derived subtree structures.
    \item {\bf GNN}: \citet{ji-etal-2019-graph} used graph neural networks (GNN) to learn token representations for graph-based dependency parsing.
    \item {\bf MP2O}: \citet{wang-tu-2020-second} used message passing to integrate second-order information to biaffine backbone.
    \item {\bf CVT}: \citet{clark-etal-2018-semi} proposed Cross-View Training, a semi-supervised approach to improve model performance.
   \item {\bf LRPTR}: \newcite{fernandez2019left} also took advantage of pointer networks to implement transition-based parser, which contains only $n$ actions and is more efficient than StackPTR.
   \item {\bf HiePTR}: \newcite{fernandez2021dependency}  introduced structural knowledge to the sequential decoding of the left-to-right dependency parser with Pointer Networks.
    \item	{\bf TreeCRF}: \newcite{zhang-etal-2020-efficient} presented a second-order TreeCRF extension to the biaffine parser.
    \item {\bf HPSG}: \citet{zhou-zhao-2019-head} used head-driven phrase structure grammar to jointly train constituency and dependency parsing.   
    \item {\bf HPSG+LA}: \citet{mrini-etal-2020-rethinking} added a label attention layer to HPSG to improve model performance. HPSG+LA also relies on the additional constituency parsing dataset.
   \item {\bf MulPTR}: \citet{fernandez2020multitask} jointly trained two separate decoders responsible for constituent parsing and dependency parsing.
    \item {\bf SynTr}: \citet{mohammadshahi2020recursive} proposed recursive non-autoregressive graph-to-graph Transformers for the iterative refinement of dependency graphs conditioned on the complete graph.
\end{itemize}

\end{document}